\documentclass[a4paper]{styles/svproc}
\usepackage{url}
\usepackage{subfigure}
\usepackage{graphicx}
\usepackage{amssymb}
\usepackage{mathrsfs}
\usepackage{algpseudocode}
\usepackage{caption}
\usepackage{amsmath}
\usepackage{algorithmicx}
\usepackage{algorithm} 
\usepackage[english]{babel}
\usepackage[autostyle, english = american]{csquotes}
\MakeOuterQuote{"}

\begin{document}
\mainmatter              % start of a contribution
\title{EcoBoat: Design and Experimental Validation of an Autonomous Body-Board Boat For Cleaning Water Bodies}
\titlerunning{EcoBoat}  % abbreviated title (for running head)
\author{M. Aman Ansari$^\dagger$ \and Saifullah Khan$^\dagger$ \and Rahul Kulkarni$^\dagger$ \and P.B. Sujit}
\authorrunning{Ansari et al.} % abbreviated author list (for running head)
\tocauthor{M. Aman Ansari, Saifullah Khan, Rahul Kulkarni, PB Sujit}

\institute{Indian Institute of Science Education and Research Bhopal, India \\
\email{ \{amanansari, mohammad21, rahulv21, sujit\}@iiserb.ac.in},\\ 
% \text{$^*$ all authors contributed equally}
% \email{@iiserb.ac.in}, \hspace{1cm}
% \email{@iiserb.ac.in},\\ 
% Website: \texttt{https://moonlab.iiserb.ac.in/}}
% \and
$^\dagger$ all authors contributed equally}

\maketitle              % typeset the title of the contribution

% \begin{abstract}
% - PUT IT AS A COVERAGE PROBLEM \\
% - THERE WILL BE A TIME LIMIT \\
% - COMPARISON BETWEEN MANUAL LABOUR AND BODY BOAT CLEANING (AMOUNT OF GARBAGE) \\
% - GARBAGE RETENTION AS A METRIC \\
% - DISCRETE OR CONTINUOUS \\
% - THE OVERALL COST OF THE BOAT IS LOW BECAUSE OF THE LOW COST USED SENSORS AND COMPONENTS \\

% We propose the design of a net structure/frame attached to an autonomous body boat to clean artificial water bodies. We developed random path planning algorithm to drive the body boat with constant speed.

%\keywords{autonomous surface vehicles, random path planning, computational geometry, cleaning water bodies}

%

\section{Introduction}
%

% One of the primary challenges associated with swimming pools is maintaining cleanliness. The water must be cleaned daily, and the cleanliness has to be maintained within certain bounds. 
Physical cleaning of the water bodies requires significant human effort to clean garbage, algae, debris and other foreign substances. Therefore, a robotic system is needed for swimming pool cleaning to reduce human effort and improve efficiency. \par

This paper addresses the problem of cleaning the surface of indoor as well as outdoor water bodies. The surface of water can have floating debris, garbage, leaves, algae, etc., depending on the way it is used and the environment surrounding it. Although existing methods are available for cleaning the water surface, they are labor-intensive, expensive, can impact water quality, and may result in water wastage. \par

In this paper, we introduce EcoBoat, a cost-effective Autonomous Surface Vehicle (ASV) designed for efficient debris collection. For indoor water bodies such as swimming pools, EcoBoat relies solely on ultrasonic sensors to detect boundaries and floating obstacles, navigating the water using a hybrid approach that combines a random walk method with boundary-following behaviour. For outdoor environments such as ponds and lakes, the system uses GPS to geofence a designated area and employs a random walk approach to ensure surface coverage for skimming. The algorithm used in this paper is simple and easy to deploy with less memory requirement. Most existing methods rely on sophisticated, costly, and fragile sensor systems and/or complex mapping techniques. In this work, we propose a minimalist approach to achieve effective coverage with minimal sensing and computation, accepting a potential trade-off in performance to clean the surface of water.

% \begin{figure}
%     \centering
%     \includegraphics[width=0.8\linewidth]{figures/EcoBoat.png}
%     \caption{Net Structure attached to the body boat}
%     \label{fig:enter-label}
% \end{figure}

\subsection{Related Work}
Existing solutions for cleaning water bodies are designed specifically for either swimming pools or natural water bodies like lakes and rivers, but not both. There is no hybrid solution that can effectively operate in both environments.

For swimming pools, cleaning robots generally fall into two categories: \\ skimmers/surface cleaners and floor-and-wall cleaners. These robots are available in both corded and cordless versions. Examples of cordless models include the solar-powered Dolphin Skimmi\cite{dolphin_skimmi}  and Aiper Surfer S1\cite{aiper_surfer}, while corded models include Dolphin Premier\cite{dolphin_premier} and Wybot L1 \cite{wybot}. Although effective, these robots rely on sophisticated and often expensive sensor systems, making them fragile and costly to maintain.

Lakes and rivers, on the other hand, require heavy machinery like amphibious excavators and weed harvesters to clear floating debris. These machines, while effective in open water, are not suitable for smaller, controlled environments like swimming pools.

This gap in existing solutions highlights the need for a hybrid cleaning robot, one that can autonomously remove floating debris from both swimming pools and natural water bodies. Developing such a system would provide a versatile and efficient alternative to the specialized cleaning solutions currently in use.

\section{Problem Statement}

Consider a water surface region $R \subseteq \mathbb{R}^{2}$ with dimensions of length \textit{l} and width \textit{w}. The objective is to maximize collection of floating debris, garbage, leaves or algae within a predefined time horizon $\Gamma$. EcoBoat, designed for efficient debris collection, utilizes low-cost ultrasonic sensors to detect boundaries in indoor environments such as swimming pools. For outdoor environments, it relies on GPS-based geofencing to define a designated area for skimming and collection.
\begin{itemize}
    \item \textbf{For indoor environment: } The robot employs a hybrid approach for coverage, which is boundary-following behaviour when it detects a boundary or an obstacle on one of its side and random walk approach, in which it moves in a straight line until detecting a boundary or obstacle within a predefined safe distance threshold $\mathcal{D}$ using ultrasonic sensors. Upon detection, the robot executes a random turn at a specified angle $\theta_{i}$ where $i = \{1, 2, ..., n\}$ from a list of angles $\Theta$ before resuming straight-line motion. 
    \item \textbf{For outdoor environment: } The robot employs a random walk-based approach for debris collection, where the designated cleaning area is geofenced using GPS. When the ASV comes within 1 meter ($\mathcal{G}$) of the geofence or has to avoid an obstacle, it executes a random turn at a specified angle  $\theta_{i}$ where $i = \{1, 2, ..., n\}$ from a list of angles $\Theta$ before resuming straight-line motion.
\end{itemize}
The aim is to make the system cost-effective and minimal. The primary challenge is effective coverage under sensory and computational constraints.
% \subsubsection{The General Case: Nontriviality.}

\section{Hardware Design and Algorithm}

\subsection{Design of EcoBoat}

The EcoBoat was designed and fabricated in-house, utilizing a modified body-surfing board made of expanded polypropylene (EPP) foam as its hull. To mount vital components, a frame structure composed of 45×45 mm aluminum extrusions was fabricated and securely mounted onto the hull using 3D-printed PLA+ mounts affixed to hard points on the underside of the body board. The trash collection mechanism consists of a detachable structure made from 20×20 mm aluminum extrusions, with its inner faces covered by a high-density polyethylene (HDPE) fishing net. This structure features a quick-release mechanism, for easy removal and disposal of collected debris after cleaning. The EcoBoat measures 1700 mm × 900 mm × 600 mm and has a dry weight of 15 kg, with an effective payload capacity of approximately 20 kg. The EcoBoat follows a randomized navigation algorithm \cite{random}, while collecting floating debris in its path.
\begin{figure}[!htb]
\minipage{0.45\textwidth}
  \includegraphics[width=\linewidth]{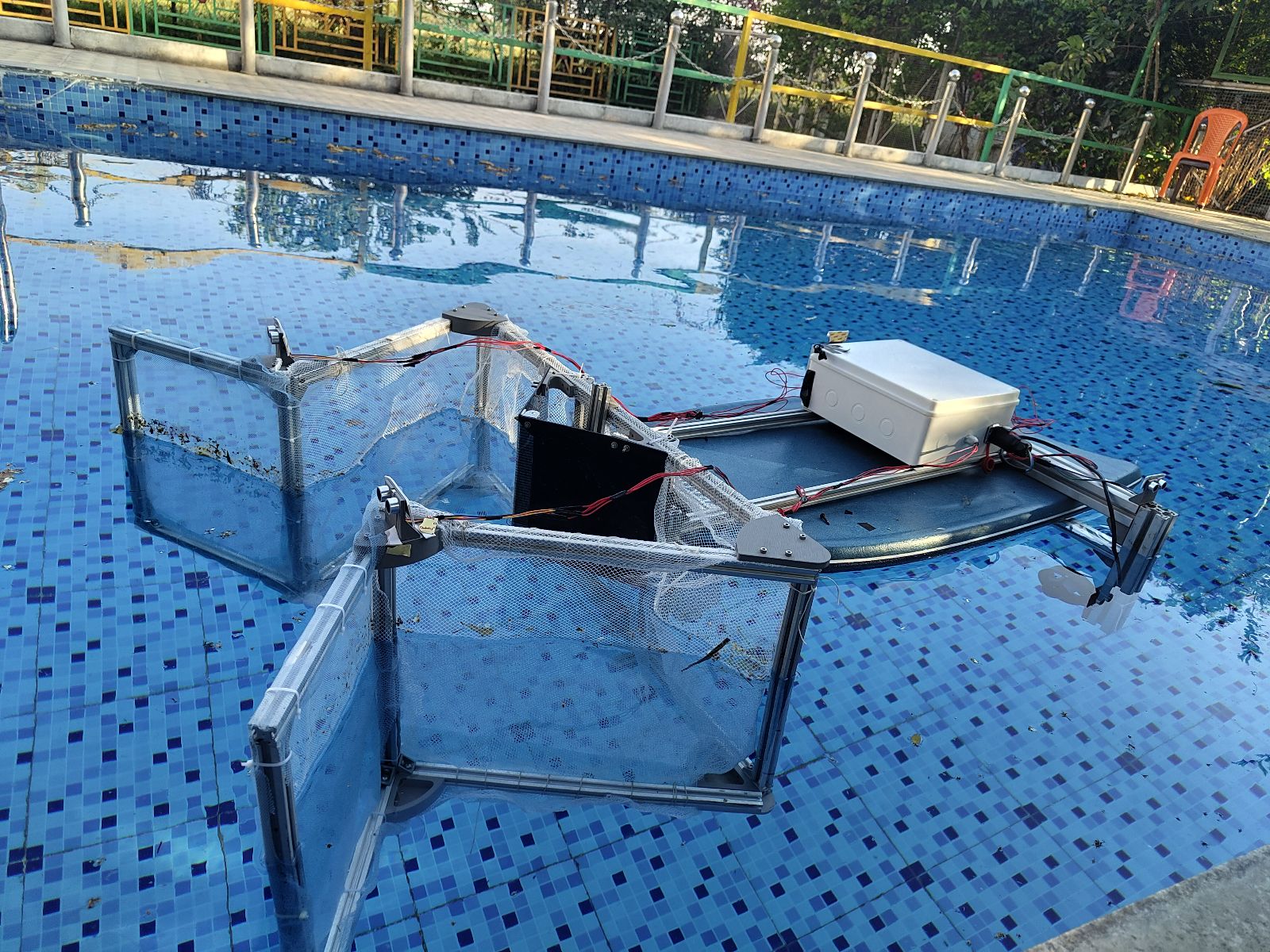}
  \caption{EcoBoat in action}\label{fig:awesome_image1}
\endminipage\hfill
\minipage{0.45\textwidth}
  \includegraphics[width=\linewidth]{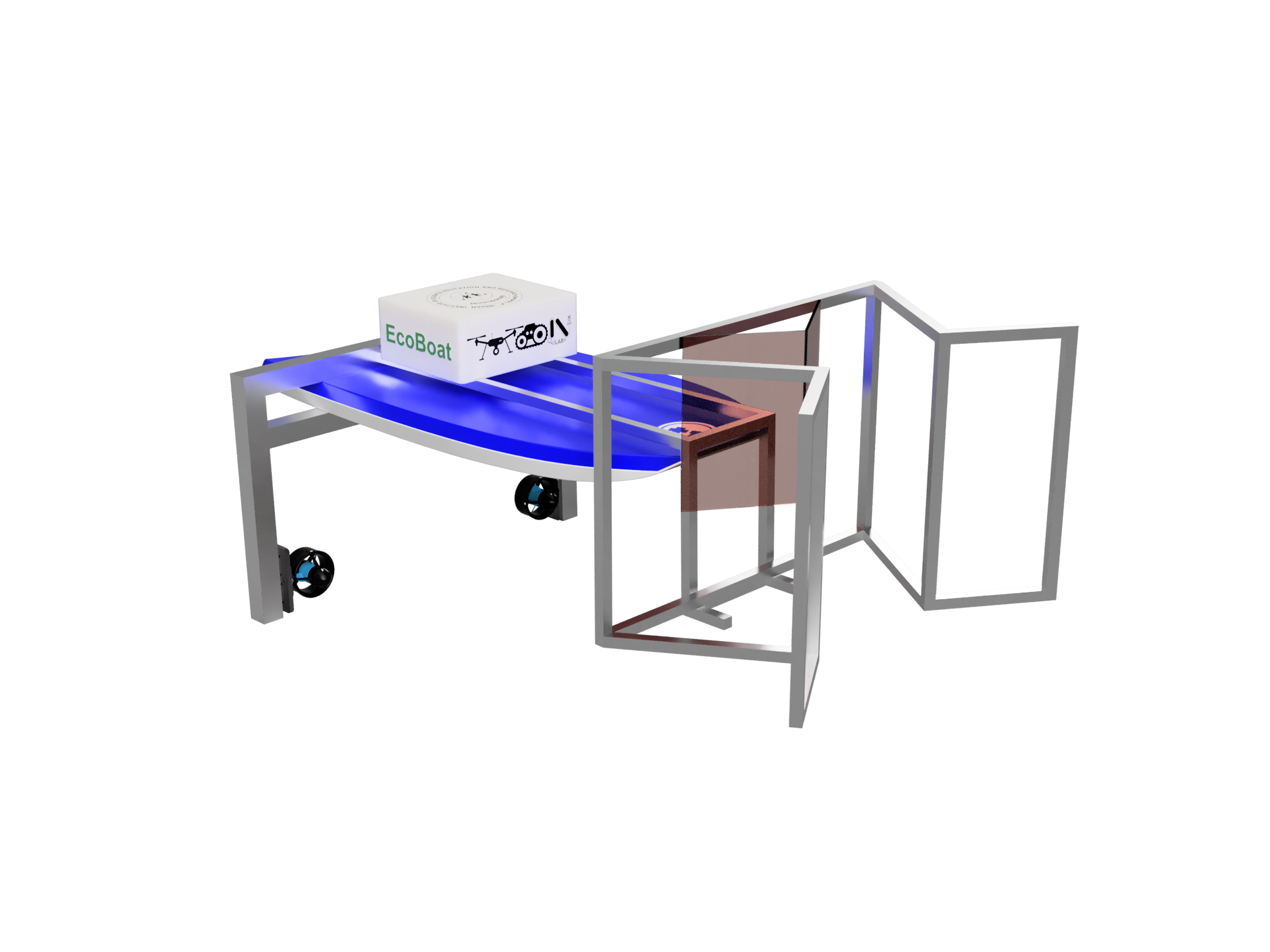}
  \caption{CAD render of EcoBoat}\label{fig:asv_cad}
\endminipage\hfill

\end{figure}

One of its key novelty is a Tesla valve-inspired \cite{tesla} collection basket, designed to allow debris intake during forward motion while preventing its escape during braking or turning maneuvers. In the EcoBoat’s design, this principle is applied using a symmetric tesla valve inspired geometry to retain collected debris efficiently. 
% As the EcoBoat moves forward, the incoming water flow carries debris into the collection basket. The internal geometry of the basket channels this flow at an acute angle against the oncoming primary flow, creating counteracting fluid forces that nullify outward movement of the debris. This interaction generates localized whirlpools within the collection area, ensuring that debris remains trapped. During braking or turning, these whirlpools try to maintain their rotational and translational inertia, gradually dissipating against the inner surfaces of the collection structure, resulting in debris getting stuck and preventing from escaping. This novelty simplifies construction, reduces costs, and enables easy disposal without requiring complex mechanisms or expensive actuators.

% A Tesla valve, originally proposed in 1920 by Nikola Tesla [cite patent], is a passive fluid diode that allows fluid to flow preferentially in one direction while restricting flow in the opposite direction.

\subsection{System Architecture}
\begin{figure}[]
    \centering
    \includegraphics[width=0.8\linewidth]{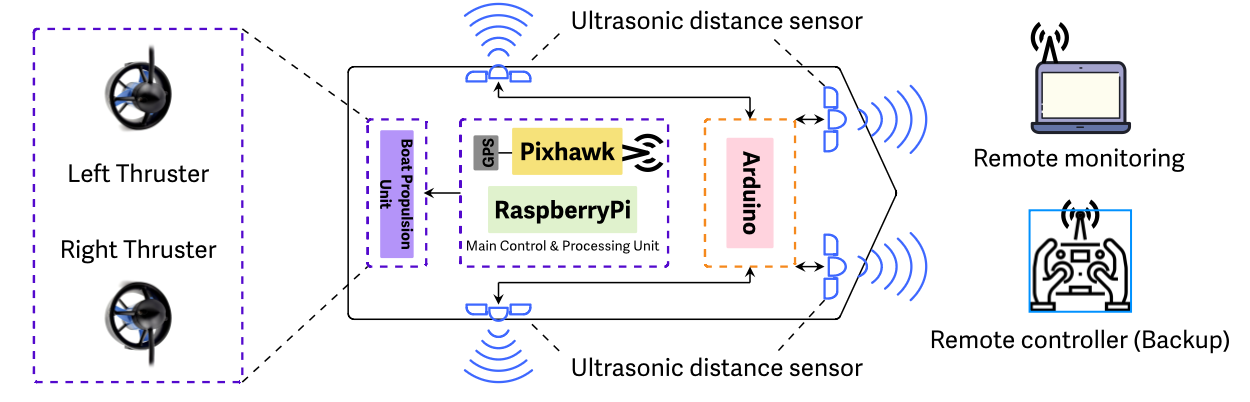}
    \caption{System architecture}
    \label{fig:arch}
\end{figure}

The boat's architecture is composed of interconnected components, as shown in Fig \ref{fig:arch}. It includes a main control system, two T200 thrusters, and a series of Li-ion batteries for independent power to each sub-system. These components drive the boat's movement based on commands from the primary processing unit, a Raspberry Pi \cite{may:ehr:stein}.  
A 2.4.8 PixHawk \cite{fost:kes} serves as the main thruster controller and is connected to GPS. Additionally, four ultrasonic distance sensors—two at the front and one on each side are used to detect obstacles and pool boundaries as well as obstacles. Their data is processed by an Arduino Mega, which transmits relevant information to the primary processing system.  
The boat also features onboard radio telemetry for remote monitoring, with a backup option for manual RC control when required.
\subsection{Boundary Detection and Obstacle Avoidance Algorithm}

EcoBoat operates in both indoor and outdoor environments, employing distinct strategies for efficient debris skimming.

\subsection{Indoor Operation}

EcoBoat employs a sensor-based obstacle avoidance and boundary detection hybrid algorithm. It relies on four ultrasonic sensors positioned at the front-right ($d_{FR}$), front-left ($d_{FL}$), right ($d_{R}$), and left ($d_{L}$) to detect obstacles and determine appropriate maneuvers.

\textbf{Key Parameters:}
$\mathcal{D}$: Safe distance threshold,  $\mathcal{Q} = \{d_{\text{FR}}, d_{\text{FL}}, d_{\text{R}}, d_{\text{L}}\}$: Distance sensor readings, $\Theta = \{\theta_1, \theta_2, \ldots, \theta_n\}$: Set of rotation angles, $\psi$: Current yaw angle,  $\psi_{\text{target}}$: Target yaw angle after rotation

The algorithm categorizes boundary or obstacle-hit scenarios into different types:
\[
\text{Hit} =
\begin{cases}
    \text{FRONT\_BOTH} & \text{if } d_{\text{FR}}, d_{\text{FL}} < \mathcal{D} \\
    \text{RIGHT\_SIDE} & \text{if } d_{\text{R}} < \mathcal{D} \\
    \text{LEFT\_SIDE} & \text{if } d_{\text{L}} < \mathcal{D} \\
    \text{FRONT\_SIDE\_RIGHT} & \text{if } d_{\text{FR}} < \mathcal{D}, d_{\text{R}} < \mathcal{D} \\
    \text{FRONT\_SIDE\_LEFT} & \text{if } d_{\text{FL}} < \mathcal{D}, d_{\text{L}} < \mathcal{D} \\
    \text{RIGHT\_FRONT\_ONLY} & \text{if } d_{\text{FR}} < \mathcal{D} \\
    \text{LEFT\_FRONT\_ONLY} & \text{if } d_{\text{FL}} < \mathcal{D} \\
\end{cases}
\]

When an obstacle is detected, EcoBoat adjusts its yaw angle using an Inertial Measurement Unit (IMU):

\[
\psi_{\text{target}} = 
\begin{cases}
    (\psi_{\text{current}} + \theta) \mod 360^{\circ}, & \text{if turning clockwise} \\
    (\psi_{\text{current}} - \theta) \mod 360^{\circ}, & \text{if turning counterclockwise}
\end{cases}
\]

Rotation continues until:

\[
|\psi_{\text{current}} - \psi_{\text{target}}| \leq 10^{\circ}
\]

Additionally, EcoBoat follows walls when boundaries are detected on either side. If no obstacles or boundaries are detected, it moves forward.

\subsection{Outdoor Operation}

EcoBoat operates within a geofenced area defined by GPS, using ultrasonic sensors to detect and avoid obstacles.

\textbf{Key Parameters:} $\mathcal{G}$: Distance threshold (1 meter), $\mathcal{Q}, \Theta, \psi, \psi_{\text{target}}$: As defined in indoor operation.

\textbf{Obstacle Classification:} (Same as in indoor operation)

To prevent geofence breaches, EcoBoat recalculates its yaw angle using the IMU and rotates until:$| \psi_{\text{current}} - \psi_{\text{target}}| \leq 10^{\circ}$. 
If no obstacles are detected and the geofence is maintained, EcoBoat continues forward.

\section{Experiments}

\subsection{Design Experiments}
We iteratively optimized our EcoBoat design through field experiments conducted in both indoor pools and lakes. These experiments provided valuable insights that guided improvements in the system’s design. Initially, we explored using a modified kayak as the hull for our autonomous surface vehicle (ASV), with a net-based collection structure attached. However, this setup proved too bulky for indoor operations. To address this, we transitioned to a more compact and lightweight bodyboard as the hull.

\begin{figure}
    \centering
    \subfigure[]{\includegraphics[width=2.9cm,height=3cm]{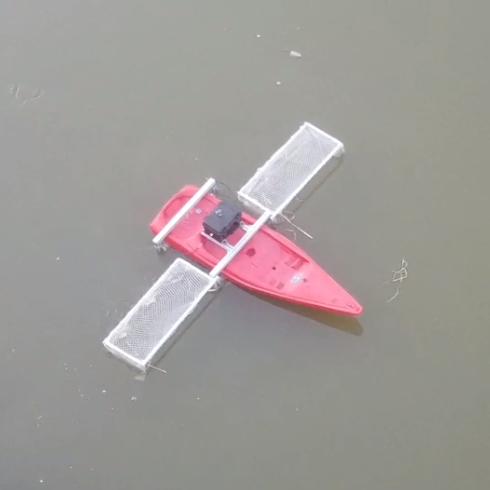}}
    \subfigure[]{\includegraphics[width=2.9cm,height=3cm]{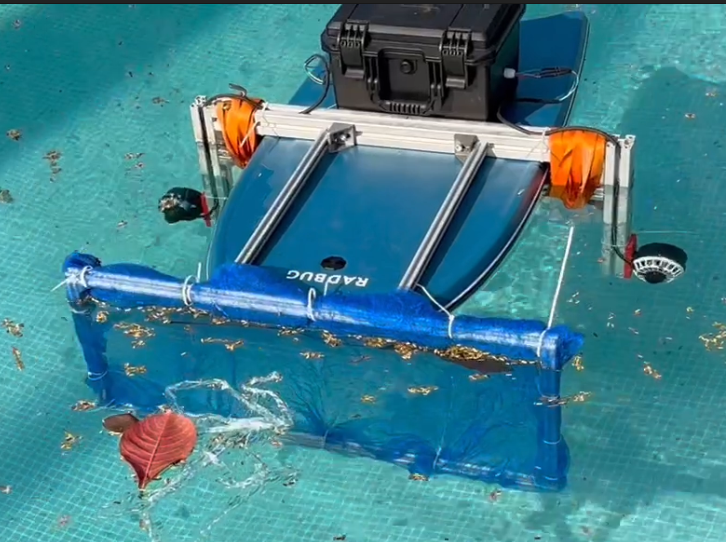}}
    \subfigure[]{\includegraphics[width=2.9cm,height=3cm]{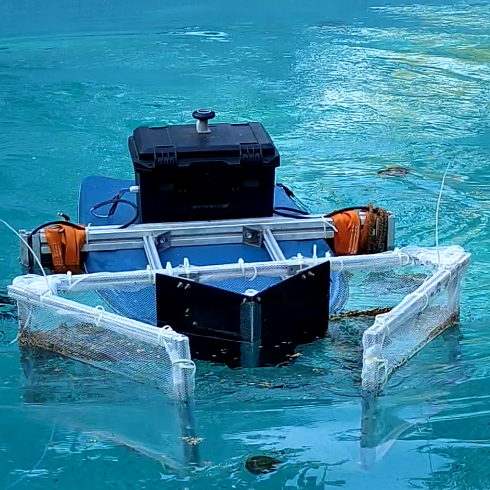}}
    \subfigure[]{\includegraphics[width=2.9cm,height=3cm]{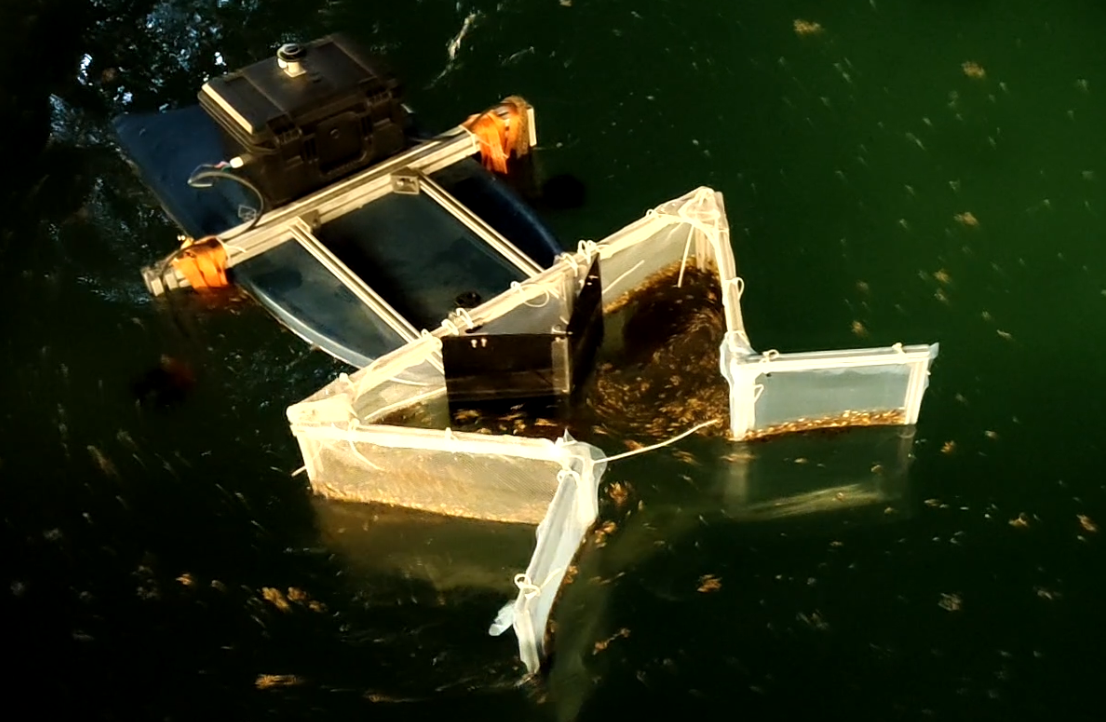}}
    \caption{(a) First Iteration (using Kayak hull) (b) Second Iteration (Shift to Bodyboard) (c) Third Iteration (Improvement in Collector Design) (d)Fourth Iteration (Best Performing Design) }
    \label{fig:four_images}
\end{figure}

Following this modification, we iteratively refined our design, testing various configurations for the collection structure. Among these, the Tesla-valve-inspired collection system demonstrated the best performance. This design enhancement significantly improved debris collection efficiency while also reducing weight and bulkiness. These improvements led to lower energy consumption and enhanced usability in both indoor and outdoor environments.

% \subsection{Experiments Scheduled}
% Experiments are scheduled (April-May) to evaluate the ASV's performance in skimming the surface of the water of a lake autonomously. The test will be conducted at Lower Lake, Bhopal, a water body with a significant amount of floating debris. The scheduled experiments also include testing the autonomous capabilities of EcoBoat in swimming pools of different dimensions and shapes, both with and without floating obstacles. We will be submitting our experimental results showing autonomous capabilities with the supplementary video as the experiment couldn't be completed by the submission deadline. 

\begin{figure}
    \centering
    \subfigure[]{\includegraphics[width=2.9cm,height=3cm]{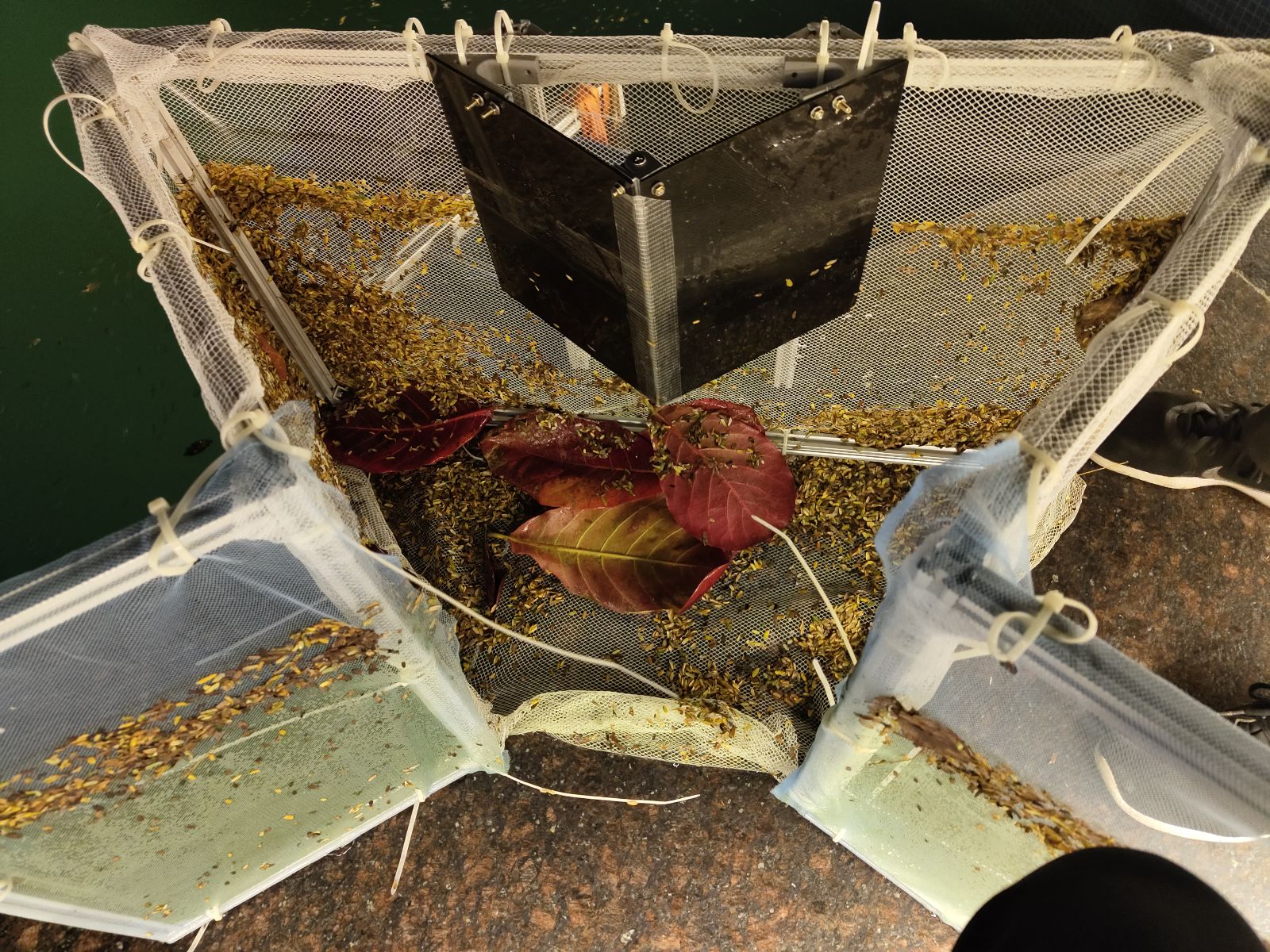}}
    \subfigure[]{\includegraphics[width=2.9cm,height=3cm]{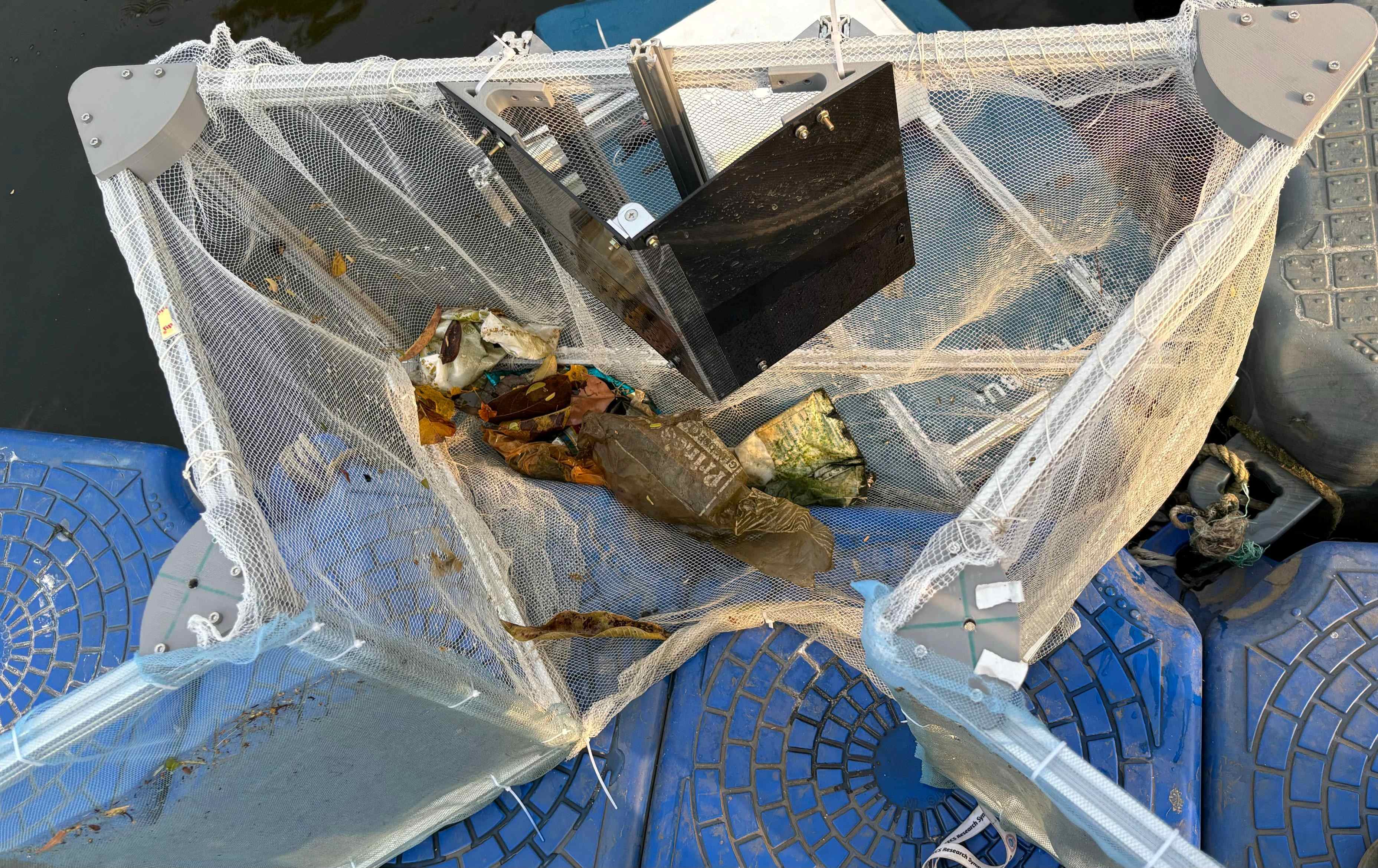}}
    \subfigure[]{\includegraphics[width=2.9cm,height=3cm]{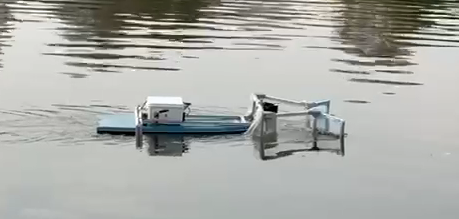}}
    \caption{(a) Garbage collected at pool (b) Garbage collected at Lake (b) Ecoboat skimming at Lake}
    \label{fig:four_images}
\end{figure}

% \section{Experimental Insights}
% LOWER LAKE PATH PHOTO. REASON WHY TRAJECTORY LOOKS LIKE THAT.

% \section{Lesson Learned}
% SHOW CHANGE IN DESIGN OF ASV OVER TIME

% \paragraph{Notes and Comments.}


\end{document}